# Boosting Adversarial Attacks on Neural Networks with Better Optimizer

Heng Yin[†], Hengwei Zhang[†], Jindong Wang and Ruiyu Dou

State Key Laboratory of Mathematical Engineering and Advanced Computing, Zhengzhou, Henan 450001, China

†These authors contributed equally to this work

Correspondence should be addressed to Hengwei Zhang; wlby_zzmy_henan@163.com

**Abstract:** Convolutional neural networks have outperformed humans in image recognition tasks, but they remain vulnerable to attacks from adversarial examples. Since these data are crafted by adding imperceptible noise to normal images, their existence poses potential security threats to deep learning systems. Sophisticated adversarial examples with strong attack performance can also be used as a tool to evaluate the robustness of a model. However, the success rate of adversarial attacks can be further improved in black-box environments. Therefore, this study combines a modified Adam gradient descent algorithm with the iterative gradient-based attack method. The proposed Adam Iterative Fast Gradient Method is then used to improve the transferability of adversarial examples. Extensive experiments on ImageNet showed that the proposed method offers a higher attack success rate than existing iterative methods. By extending our method, we achieved a state-of-the-art attack success rate of 95.0% on defense models.

## 1. Introduction

In image recognition tasks, convolutional neural networks are able to classify images with an accuracy approaching that of humans [1-4]. However, researchers have found that neural networks are also vulnerable to adversarial examples. Szegedy et al. [5] first proposed the concept of adversarial examples: images added with small perturbations, which cause neural network models to output incorrect classifications with high confidence. These adversarial perturbations are often indistinguishable to the human eyes (In other words, there is no obvious visual difference between the adversarial examples and the original images).

Adversarial attacks can be categorized into white-box and black-box attacks. A variety of techniques can be used to generate adversarial examples and perform white-box attacks, depending on the model structure and corresponding parameters [6-10]. In addition, adversarial examples are generally transferable as data generated for one model may be able to fool other models. This facilitates black-box attacks, in which the structure and parameters of the model are not available, for various neural networks [11]. Goodfellow et al. [6] suggested that different models learn similar decision boundaries during the same image classification tasks and obtain similar parameters. These properties make it easier to generalize adversarial examples to different models.

Although adversarial examples are generally transferable, the optimal approach of improving this transferability needs to be examined further. Due to a balance between attack performance and transferability, basic iterative attacks are often more effective than single-step attacks in white-box environments and weaker in black-box environments. In white-box attacks, iterative methods excessively fit specific network parameters, achieving a high success rate but preventing generalization to other models [9]. We consider that this is the result of overfitting, since attack performance for adversarial examples in white-box and black-box environments is similar to the neural network performance on training and test sets.

Unlike white-box attacks, black-box attacks are more consistent with actual attack-defense environments and are primarily performed in one of three ways. 1) Decision-based attacks are conducted using information collected from the network. Although access to model structure and parameters is unavailable, the attacker can generate adversarial examples by executing multiple queries against the model [12-15]. 2) In substitute-model techniques, the attacker can input images into the model to obtain an output label. Then, the substitute model can be trained to imitate the target model and generate adversarial examples [16]. 3) Transferability attacks are conducted by improving the transferability of adversarial examples generated in a white-box setting [8, 9]. The first two black-box attacks require large quantities of model queries, which is impractical in some cases (e.g., online platforms often limit the number of queries). As such, this study investigates black-box attacks using adversarial data with strong transferability. We investigate a state-of-the-art optimizer Adam for improving black-box adversarial attacks.

In this paper, we propose the Adam Iterative Fast Gradient Method (AI-FGM) to improve the transferability of adversarial examples among different models [17]. Inspired by the fact that Adam is better than momentum in op-timization, we adapt Adam optimizer into the iterative gradient-based attack, so as to accelerate data update in dimensions with small gradients and achieve better convergence, by applying the second momentum term and a decreasing step size. As shown in Figure. 1, different from the momentum method which just accumulates the gradients of data points along the optimization path, our Adam method can also accumulate the square of the gradients. Such accumulation might help obtain an adaptive update direction, which leads to better local minimum. Besides, a variable step size is also used to avoid oscillating.

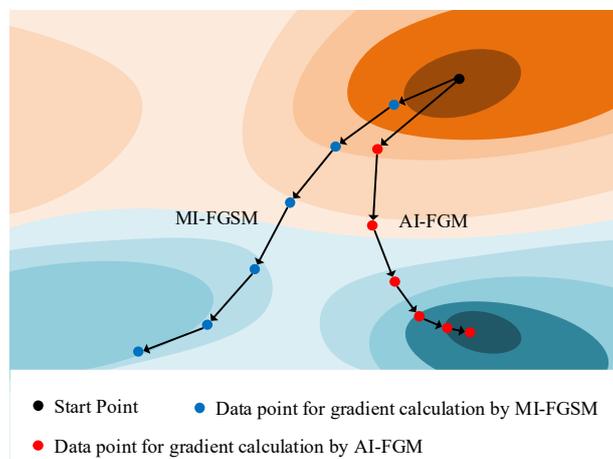

**Figure 1.** Schematic optimization path of MI-FGSM and the proposed AI-FGM. MI-FGSM accumulates the gradients of data points along the optimization path, while Adam accumulates the gradients and the square of the gradients. MI-FGSM adopts an invariable step size, while AI-FGM adopts a decayed step size that leads to better convergence.

Compared with existing gradient-based methods, the proposed method offers improved attack performance in black-box settings. This approach was tested on multiple networks, including adversarially trained networks, achieving higher attack success rates on black-box models. In addition, we combined our approach with advanced methods and attacked an ensemble of multiple networks, which further improved the transferability of adversarial examples.

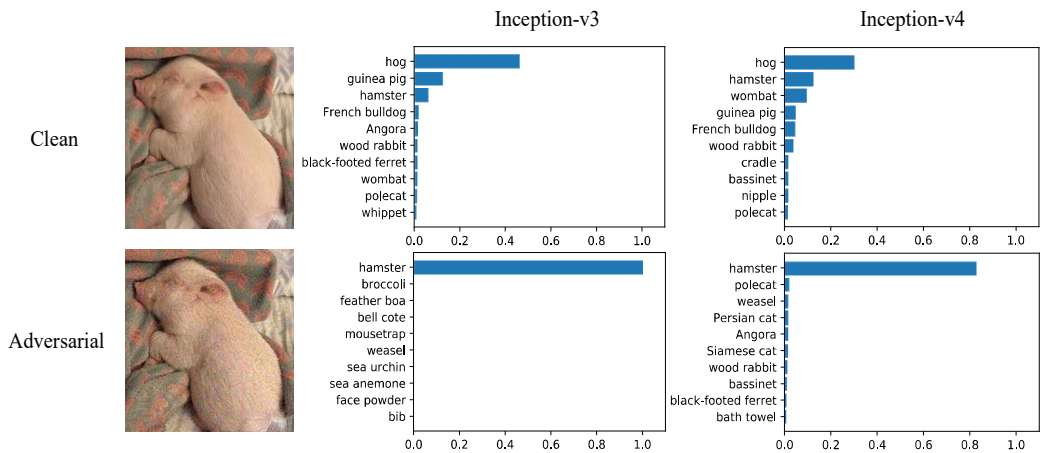

**Figure 2.** Classification of a normal image and corresponding adversarial example by Inc-v3 and Inc-v4. The first row shows the top-10 confidence distributions for a clean image, for which both models provided a correct prediction. The second row shows the top-10 confidence distributions for the adversarial example generated for Inc-v3 by AI-FGM. It is evident that the adversarial example successfully attacked Inc-v3 (white-box) and Inc-v4 (black-box) with high confidence.

## 2. Related Work

### 2.1. Adversarial Attack Methods

A variety of techniques have been proposed to generate transferable adversarial examples, which can be divided into two categories: optimizer-based methods and data-augmentation-based methods. Specifically, Goodfellow et al. proposed the Fast Gradient Sign Method (FGSM) to generate adversarial examples with very low calculation costs [6]. By extending FGSM, Alexey et al. developed the Iterative Fast Gradient Sign Method (I-FGSM)

[7]. Dong et al. proposed the Momentum Iterative Fast Gradient Sign Method (MI-FGSM) to improve the transferability of adversarial examples in black-box environments [8]. Lin et al. proposed Nesterov Iterative Fast Gradient Sign Method (NI-FGSM) and adapted Nesterov accelerated gradient into the iterative attacks [18]. Xie et al. first applied data augmentation into the generation of adversarial examples and proposed Diverse Inputs Method (DIM) [9]. Dong et al. proposed Translation-Invariant attack (TI-FGSM) and improved the transferability of adversarial examples on defense models by a large margin [19]. Lin et al. proposed Scale-Invariant attack method (SIM) and combined it with existing methods, resulting SI-NI-TI-DIM (the currently strongest black-box attack method) [18].

### 2.2. Adversarial Defense Methods

Multiple defense mechanisms have been proposed to protect deep learning models from the threat of adversarial examples [20-29]. Among these, adversarial training is the most effective way to improve model robustness [6, 30]. In this process, adversarial examples are generated and added to the training set to participate in the model training procedure. Since normal adversarial training is still vulnerable to adversarial examples, Tramèr et al. proposed ensemble adversarial training to further improve robustness [31]. In this process, adversarial data generated by multiple models are included in the training set for a single model, thereby producing a more robust classifier.

## 3. Methodology

This section provides a detailed introduction to our proposed methodology. Let $x$ denote an input image and $y$ denote the corresponding ground-truth label. The term $\theta$ represents network parameters and $J(\theta, x, y)$ describes a loss function, typically the cross-entropy loss. The primary objective is to generate an adversarial example $x^*$ to fool the model by maximizing $J(\theta, x, y)$, such that the prediction label $y^{pre} \neq y$. In this paper, we used an $L_\infty$ norm bound to limit adversarial perturbations, such that $\|x^* - x\|_\infty \leq \varepsilon$, where $\varepsilon$ refers to the size of the perturbation. The generation of adversarial examples can therefore be converted into the following optimization problem:

$$\arg\max_{x^*} J(\theta, x^*, y), \quad s.t. \|x^* - x\|_\infty \leq \varepsilon \tag{1}$$

### 3.1. Attack Methods Based on Gradient

In this section, several techniques that are used to solve the above optimization problem are introduced briefly.
- **Fast Gradient Sign Method (FGSM) [6]**: As one of the simplest techniques, it seeks adversarial examples in the direction of the gradient of the loss function with respect to the input image $x$. The method can be expressed as:

$$x^* = x + \varepsilon \cdot sign(\nabla_x J(\theta, x, y)) \tag{2}$$

- **Iterative Fast Gradient Sign Method (I-FGSM) [7]**: This algorithm is an iterative version of FGSM. The approach involves dividing the FGSM gradient operation into multiple steps that can be expressed as follows:

$$x_0^* = x, \quad x_{i+1}^* = x_i^* + \alpha \cdot sign(\nabla_x J(\theta, x_i^*, y)), \tag{3}$$

$$x_{i+1}^* = Clip(x_{i+1}^*, x - \varepsilon, x + \varepsilon), \tag{4}$$

where $\alpha$ denotes the step size of each iteration and $\alpha = \varepsilon / T$, in which $T$ denotes the number of iterations. The $Clip$ function was included to limit the adversarial example $x^*$ within the $\varepsilon$ neighborhood of the original image $x$ and satisfy the $L_\infty$ norm constraint. I-FGSM is more effective than FGSM in white-box environments, but less effective in black-box environments. In other words, I-FGSM exhibits poor transferability. This iterative attack method is also known as Projected Gradient Descent (PGD) if the algorithm is added by a random initialization on $x$ [20].
- **Momentum Iterative Fast Gradient Sign Method (MI-FGSM) [8]**: In this method, a momentum term is applied to the iterative process to escape from poor local maxima. This produces adversarial examples with more transferability, which can be expressed as [8]:

$$x_0^* = x, g_0 = 0, g_{i+1} = \mu \cdot g_i + \frac{\nabla_x J(\theta, x_i^*, y)}{\|\nabla_x J(\theta, x_i^*, y)\|_1}, \tag{5}$$

$$x_{i+1}^* = x_i^* + \alpha \cdot sign(g_{i+1}), \tag{6}$$

$$x_{i+1}^* = Clip(x_{i+1}^*, x-\varepsilon, x+\varepsilon), \tag{7}$$

where $\mu$ denotes the decay factor of the momentum term and $g_i$ denotes the weighted accumulation of gradients in the first $i$ rounds of iterations.

## 3.2. Adam Iterative Fast Gradient Method

The generation of adversarial examples is similar to the training of neural networks, both of the two process can be viewed as an optimization problem. Specifically, the adversarial example can be viewed as the training parameter, while the white-box model can be viewed as the training set and the black-box model can be viewed as testing set. From this point of view, the transferability of the adversarial examples is similar with the generalization of the models. Therefore, we can apply the methods used to improve the generalization of the models to the generation of adversarial examples. There are many methods proposed to improve the generalization of neural network models, which can be divide into two categories: better optimization and data augmentation. Correspondingly, these two kinds of methods can be used into adversarial attack, and there have been attempts to do so, e.g., MI-FGSM and DIM [8][9]. Based on the analysis above, we aim to improve the transferability of adversarial examples with Adam optimizer since it performs well in the training of neural networks.

Adam [17] is an optimization algorithm combining momentum [32] and RMSProp [33]. In the iteration process, Adam accumulates not only the gradients of the loss function with respect to the input image, but also the square of the gradients. This is done to accelerate loss function ascent in dimensions with small gradients. In addition, Adam uses a decreasing step size in each iteration to achieve better convergence.

---

**Algorithm 1** Adam Iterative Fast Gradient Method

**Input:** A convolutional neural network $f$ and the corresponding cross-entropy loss function $J(\theta, x, y)$; an original image $x$ and the corresponding ground-truth label $y$; the number of iterations $T$; the iteration time step $t$; the dimension of the input image $N$; the size of the perturbation $\varepsilon$; Adam decay factors $\beta_1$ and $\beta_2$; and a denominator stability factor $\delta$.

**Output:** An adversarial example $x^*$, s.t. $\|x-x^*\|_\infty \leq \varepsilon$.

1: $m_0 = 0$, $v_0 = 0$, $x_0^* = x$, and $t = 0$;
2: $\alpha = \varepsilon \cdot \sqrt{N}$; (8)
3: **while** $t < T$ **do:**
4: $\quad g_t = \dfrac{\nabla_x J(\theta, x_t^*, y)}{\|\nabla_x J(\theta, x_t^*, y)\|_1}$; (9)
5: $\quad m_{t+1} = \beta_1 m_t + (1-\beta_1) g_t$; (10)
6: $\quad v_{t+1} = \beta_2 \cdot v_t + (1-\beta_2) \cdot g_t^2$; (11)
7: $\quad s_{t+1} = \dfrac{m_{t+1}}{\delta + \sqrt{v_{t+1}}}$; (12)
8: $\quad \alpha_t = \alpha \cdot \dfrac{\sqrt{1-\beta_2^{t+1}}}{1-\beta_1^{t+1}} \bigg/ \sum_{i=0}^{T-1} \dfrac{\sqrt{1-\beta_2^{i+1}}}{1-\beta_1^{i+1}}$; (13)
9: $\quad x_{t+1}^* = x_t^* + \alpha_t \cdot \dfrac{s_{t+1}}{\|s_{t+1}\|_2}$; (14)
10: $\quad x_{t+1}^* = Clip(x_{t+1}^*, x-\varepsilon, x+\varepsilon)$; (15)
11: $\quad t = t+1$;
12: **end while**
13: **return** $x^* = x_T^*$.

In contrast, both I-FGSM and MI-FGSM adopt a constant step size. In the latter stages of the algorithm, oscillations occur near the local maximum, which do not converge well. To solve this problem, we have improved the Adam algorithm by normalizing the step size sequence (defined in Eq. (13)) and have applied it to the iterative gradient method. The proposed Adam Iterative Fast Gradient Method (AI-FGM) is summarized in Algorithm 1.

Specifically, the gradient $\nabla_x J(\theta, x_t^*, y)$ in each iteration is normalized by its own $L_1$ distance, defined in Eq. (9), because the scale of these gradients differs widely in each iteration [8]. Similar to MI-FGSM, $m_t$ accumulates gradients of the first $t$ iterations with a decay factor $\beta_1$, defined in Eq. (10). The result can be considered the first momentum. The term $v_t$ represents the second momentum, which accumulates the squares of gradients for the first $t$ iterations with a decay factor $\beta_2$, defined in Eq. (11). It is noteworthy that $g_t^2$ denotes an elementwise square $g_t \odot g_t$, where $\odot$ represents the Hadamard product [32]. The terms $\beta_1$ and $\beta_2$ are typically defined in the range $(0,1)$. The update direction of input $x$ is defined in Eq. (12), where the stability coefficient $\delta$ is set to avoid a zero in the denominator. The $s_t$ term is advantageous as it prompts $x$ to escape from local maxima and accelerates the updating of $x$ in dimensions with small gradients.

Learning rate decay is often used in the training of neural networks. In this study, the step size used in the Adam optimizer is continually reduced to help improve the convergence of the algorithm. If $\beta_1$ and $\beta_2$ are set appropriately, the value of $\sqrt{1-\beta_2^t}/(1-\beta_1^t)$ will decrease with the increase of $t$. Thus, a decreasing sequence can be generated when $t$ ranges from $1$ to $T$. The sequence can then be normalized to obtain the weight of step sizes in each iteration, relative to the total step size $\alpha$. This can be used to acquire a set of exponential decay step sizes controlled by $\beta_1$ and $\beta_2$, as defined in Eq. (13).

Existing techniques typically use the sign function to satisfy the $L_\infty$ norm limitation. However, if the sign function was applied in Eq. (14), the update direction in our method would be equivalent to that of MI-FGSM. Hence, we constrain adversarial examples within the $L_\infty$ norm bound by the *Clip* function and apply the step size and update direction within the corresponding $L_2$ norm bound, defined in Eqs. (14) and (15). The relationship between the $L_\infty$ norm bound ($\varepsilon$) and the $L_2$ norm bound ($\alpha$) is defined in Eq. (8), where $N$ represents the dimension of the input image $x$.

### 3.3. Attacking an Ensemble of Networks

The proposed method was also used to attack an ensemble of networks. If an adversarial example poses a threat to multiple networks, it is far more likely to transfer to other models [11]. We followed the ensemble strategy proposed by Dong et al., in which multiple models are attacked by fusing network logits [8]. Specifically, in order to attack an ensemble of $K$ models, logits were fused as follows:

$$l(x) = \sum_{k=1}^{K} \omega_k l_k(x), \tag{16}$$

where $l_k(x)$ refers to the logit output of the kth model, $\omega_k$ refers to the ensemble weight ($\omega_k \geq 0$), and $\sum_{k=1}^{K} \omega_k = 1$.

### 4. Experiment

Extensive validation experiments were conducted to demonstrate the effectiveness of the proposed methodology. Experimental settings are provided in Sec. 4.1. The results of attacking a single network and some parameters that influence the results are discussed in Sec. 4.2. Thus, results of attacking an ensemble of models are showed in Sec. 4.3. Finally, Sec. 4.4 exhibits the results of attacks by combination of AI-FGM and existing methods[1].

### 4.1. Experimental Setup

- **Dataset**: If the original images cannot be classified correctly by the network, it is meaningless to generate adversarial examples based on these data. Therefore, we selected 1000 images belonging to the 1000 categories from ImageNet validation set, all of which could be classified correctly by the tested networks. All images were scaled to $299 \times 299 \times 3$. As such, $N = 299 \times 299 \times 3$ in Sec. 3.2.

---

[1] Source code is available at https://github.com/YinHeng121/Adam_attack

- **Networks**: Seven different models were studied, four of which were normally trained networks (i.e., Inception-v3 (Inc-v3) [34], Inception-v4 (Inc-v4) [35], Inception-Resnet-v2 (IncRes-v2) [35], and Resnet-v2-101 (Res-101) [36]). The other three models were adversarially trained networks (i.e., ens3-adv-Inception-v3 (Inc-v3$_{ens3}$), ens4-adv-Inception-v3 (Inc-v3$_{ens4}$), and ens-adv-Inception-ResNet-v2 (IncRes-v2$_{ens}$) [31]).
- **Other details**: The proposed method was compared to the other methods discussed in Sec. 3.1. All the experiments described in this paper were based on the $L_\infty$ norm. Thus, the momentum decay factor $\mu$ in Eq. (5) was 1.0 and the stability coefficient $\delta$ in Eq. (12) was $10^{-8}$, as suggested in [17].

**Table 1.** Attack success rates (%) for all seven networks included in the study. The diagonal blocks indicate white-box attacks, while the off-diagonal blocks indicate black-box attacks.

|  | Attack | Inc-v3 | Inc-v4 | IncRes-v2 | Res-101 | Inc-v3$_{ens3}$ | Inc-v3$_{ens4}$ | IncRes-v2$_{ens}$ |
|---|---|---|---|---|---|---|---|---|
| **Inc-v3** | FGSM | 72.2 | 32.1 | 31.7 | 32.3 | 10.3 | 10.6 | 4.2 |
|  | I-FGSM | **100.0** | 27.5 | 23.0 | 20.9 | 6.2 | 4.7 | 1.8 |
|  | PGD | 99.8 | 18.9 | 14.7 | 14.4 | 6.1 | 6.2 | 3.2 |
|  | MI-FGSM | **100.0** | 54.3 | 50.6 | 44.0 | 13.9 | 13.3 | 6.5 |
|  | AI-FGM | **100.0** | **60.7** | **55.8** | **50.2** | **17.0** | **16.8** | **8.5** |
| **Inc-v4** | FGSM | 39.4 | 64.9 | 29.5 | 33.0 | 12.2 | 11.1 | 5.0 |
|  | I-FGSM | 43.8 | **100.0** | 27.4 | 23.4 | 6.1 | 6.3 | 2.4 |
|  | PGD | 32.0 | 99.8 | 17.7 | 16.6 | 6.4 | 5.8 | 3.1 |
|  | MI-FGSM | 69.9 | **100.0** | 57.9 | 54.1 | 19.6 | 17.7 | 8.7 |
|  | AI-FGM | **72.9** | **100.0** | **60.1** | **57.1** | **21.7** | **19.5** | **10.3** |
| **IncRes-v2** | FGSM | 37.6 | 31.8 | 57.9 | 31.4 | 13.7 | 12.0 | 6.8 |
|  | I-FGSM | 46.1 | 35.0 | 99.4 | 30.4 | 7.3 | 6.7 | 4.4 |
|  | PGD | 30.1 | 23.0 | 97.3 | 18.3 | 6.1 | 5.7 | 2.7 |
|  | MI-FGSM | 73.5 | 69.3 | **99.5** | 60.0 | 27.0 | 23.0 | 16.7 |
|  | AI-FGM | **74.6** | **71.1** | **99.5** | **61.8** | **31.3** | **25.7** | **20.5** |
| **Res-101** | FGSM | 38.3 | 33.0 | 30.2 | 79.3 | 14.6 | 13.3 | 6.4 |
|  | I-FGSM | 35.1 | 28.3 | 25.1 | 99.5 | 8.4 | 6.7 | 3.7 |
|  | PGD | 30.9 | 22.4 | 20.9 | **99.6** | 7.3 | 7.2 | 3.4 |
|  | MI-FGSM | 60.0 | 55.3 | 50.6 | 99.5 | 22.9 | 19.8 | 11.3 |
|  | AI-FGM | **64.0** | **57.7** | **54.0** | 99.5 | **27.2** | **24.1** | **15.4** |

*4.2. Attacking a Single Network*

We first performed adversarial attacks on a single network with FGSM, I-FGSM, MI-FGSM, and AI-FGM. And the adversarial examples were generated on four normally trained networks and tested on all seven networks. The results are shown in Table 1, where the success rates are misclassification rates for the corresponding models, with adversarial examples used as input. The decay factors $\beta_1$ and $\beta_2$ in AI-FGM were set to 0.99 and 0.999, respectively. The maximum perturbation $\varepsilon$ was 16. The number of iterations $T$ for I-FGSM, PGD, MI-FGSM, and AI-FGM was 10. These methods are hereafter referred to as "iterative methods" without ambiguity. The effects of these parameter choices are discussed further in this section.

As shown in Table 1, all four iterative methods attacked a white-box model with a near 100% success rate. AI-FGM performed better than the other three methods on all black-box models. For example, for adversarial examples generated on Inc-v3, AI-FGM had a success rate of 60.7% on Inc-v4, while MI-FGSM, PGD, I-FGSM and FGSM reached 54.3%, 18.9%, 27.5%, and 32.1%, demonstrating the effectiveness of the proposed method. An original image and the corresponding adversarial example generated for Inc-v3 by AI-FGM are shown in Figure. 2.

4.2.1. Decay Factors

The terms $\beta_1$ and $\beta_2$ control not only the decay amplitude of the step sizes, but also the accumulation intensity of gradients for $m_t$ and $v_t$. As such, they have a direct impact on attack success rates. We applied a grid-search method to identify an optimal set of $\beta_1$ and $\beta_2$ values. In the experiments, the maximum perturbation $\varepsilon$ was set to 16 and the number of iterations $T$ was set to 10. The values of $\beta_1$ and $\beta_2$ ranged from 0 to 1. Notably, we not only chose the values of $\beta_1$ and $\beta_2$ uniformly from 0.1 to 0.9, but also selected values close to 0 and 1, as shown in Figure. 3. This was done to provide a more comprehensive study on the effects of decay factors, as we propose that the relationship between the success rate and the decay factors may be nonlinear. Adversarial examples were then generated on Inc-v3 with AI-FGM and used to perform attacks on Inc-v3, Inc-v4, and Inc-v3$_{ens4}$. It is evident from the figure that regardless of the values of $\beta_1$ and $\beta_2$, attack success rates were al-ways near 100% in white-box environments. Beyond that, attack success rates were more sensitive to $\beta_1$ for black-box models, regardless of whether networks were normally or adversarially trained. Success rates were maximized when both $\beta_1$ and $\beta_2$ were close to 1.

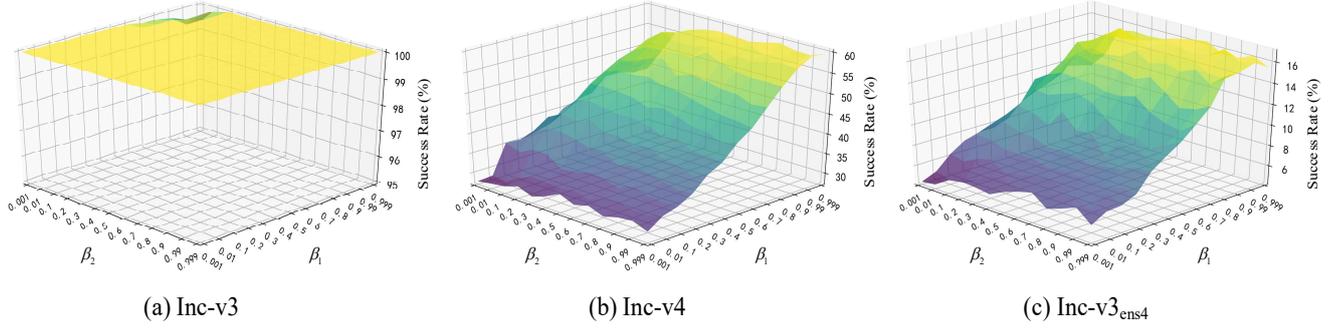

(a) Inc-v3　　　　　　　　　　(b) Inc-v4　　　　　　　　　　(c) Inc-v3$_{ens4}$

**Figure 3.** Attack success rates (%) of adversarial examples generated for Inc-v3 with AI-FGM, applied to Inc-v3 (white-box), Inc-v4 (black-box and normally trained), and Inc-v3ens4 (black-box and adversarially trained) with $\beta_1$ and $\beta_2$ in the range of $(0,1)$.

#### 4.2.2. The Number of Iterations

The effect of iteration quantities on success rates was studied by performing attacks using iterative methods. The maximum perturbation $\varepsilon$ was set to 16 and the decay factors $\beta_1$ and $\beta_2$ were set to 0.99 and 0.999, respectively. The number of iterations $T$ ranged from 1 to 20. Adversarial examples generated on Inc-v3 with I-FGSM, MI-FGSM, and AI-FGM were then used to attack Inc-v3 and Inc-v4, as shown in Figure. 4.

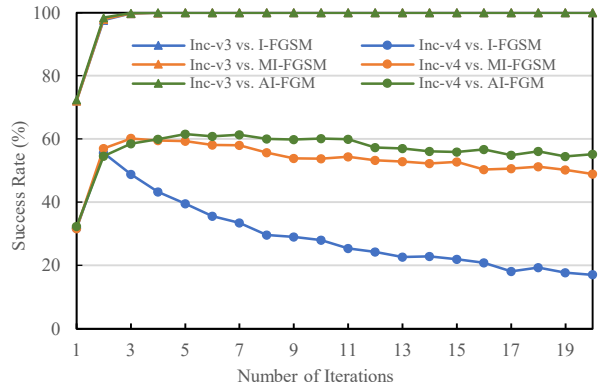

**Figure 4.** Attack success rates (%) of adversarial examples generated for Inc-v3 with I-FGSM, MI-FGSM, and AI-FGM, applied to Inc-v3 (white-box) and Inc-v4 (black-box) with $T$ ranging from 1 to 20. It is noteworthy that the curves of Inc-v3 vs. I-FGSM, Inc-v3 vs. MI-FGSM, and Inc-v3 vs. AI-FGM overlap.

As shown in figure, the success rates of black-box attacks for several iterative methods decreased with the increase of the iteration quantities. However, AI-FGM always outperformed I-FGSM and MI-FGSM for a given value of $T$.

#### 4.2.3. Perturbation Size

The effect of adversarial perturbation size on attack success rates was studied by setting the number of iterations $T$ to 10, with decay factors of 0.99 and 0.999. The size of perturbations $\varepsilon$ ranged from 1 to 30. Adversarial examples generated on Inc-v3 with I-FGSM, MI-FGSM, and AI-FGM were used to attack Inc-v3 and Inc-v4, as shown in Figure. 5.

It is evident from the figure that attack success rates can reach 100% in white-box environments. Success rates for black-box models increased steadily with increasing values of $\varepsilon$. AI-FGM always outperformed I-FGSM and MI-FGSM for a given value of $\varepsilon$. In other words, AI-FGM can achieve comparable black-box attack success rates with smaller perturbations.

### 4.3. Attacking an Ensemble of Networks

The experimental results presented above suggest that AI-FGM can increase the transferability of adversarial examples. In addition, the success rate of black-box attacks can be further improved by attacking an ensemble of networks. As discussed in Sec. 3.3, multiple models were attacked by fusing network logits. In the experiment, all seven networks described in Sec. 4.1 were used, and we generated adversarial examples on the ensemble of Inc-v3, Inc-v4, IncRes-v2, and Res-101, with FGSM, I-FGSM, MI-FGSM, NI-FGSM, and AI-FGM. Attacks were then performed on the other three defense models. Decay factors were set to 0.99 and 0.999, the number of iterations was 10, the maximum perturbation was 16, and each network had an equal ensemble weight of $\omega_k = 1/4$. The corresponding results are shown in Table 2, and it is evident that AI-FGM was more effective than the other four methods on adversarially trained models.

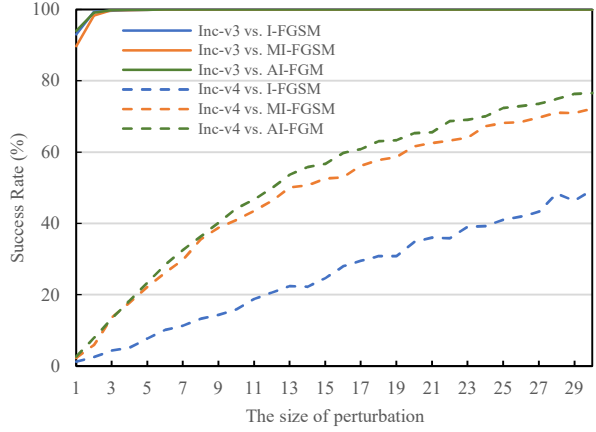

**Figure 5.** Attack success rates (%) of adversarial examples generated for Inc-v3 with I-FGSM, MI-FGSM, and AI-FGM, applied to Inc-v3 (white-box) and Inc-v4 (black-box) with $\varepsilon$ ranging from 1 to 30. It is noteworthy that the curves of Inc-v3 vs. I-FGSM, Inc-v3 vs. MI-FGSM, and Inc-v3 vs. AI-FGM overlap.

**Table 2.** Attack success rates (%) on adversarially trained networks under ensemble-model setting. The adversarial examples are generated on the ensemble of Inc-v3, Inc-v4, IncRes-v2, and Res-101.

| Attack | Inc-v3$_{ens3}$ | Inc-v3$_{ens4}$ | IncRes-v2$_{ens}$ | Average |
|---|---|---|---|---|
| FGSM | 18.6 | 17.3 | 9.5 | 15.1 |
| I-FGSM | 20.1 | 16.9 | 11.4 | 16.1 |
| MI-FGSM | 49.4 | 43.2 | 25.9 | 39.5 |
| NI-FGSM | 46.7 | 41.6 | 23.0 | 37.1 |
| AI-FGM (Ours) | **56.2** | **50.6** | **33.0** | **46.6** |

*4.4. Combination with Advanced Methods*

In this subsection, we combined our AI-FGM with SI-NI-TI, SI-NI-DI, SI-NI-TI-DI [18] respectively, and compared the black-box attack success rates of our extensions with the original methods under single-model setting. The adversarial examples were generated on Inc-v3, with the number of iterations set to 10 and the maximum perturbation to 16 respectively. It is noteworthy that our combination with other methods is more like an improvement. For example, the combination of AI-FGM and SI-NI-TI was done by replacing the Nesterov optimizer in SI-NI-TI with Adam, thus resulting SI-AI-TI. As shown in Table 3, our method SI-AI-TI-DI achieved an average attack success rate of 65.1%, surpassing the state-of-the-art attack by 10.7%.

**Table 3.** Comparison of combined methods and original methods on attack success rates (%) against adversarially trained networks under single-model setting. The adversarial examples are generated on the ensemble of Inc-v3, Inc-v4, IncRes-v2, and Res-101.

| Attack | Inc-v3$_{ens3}$ | Inc-v3$_{ens4}$ | IncRes-v2$_{ens}$ | Average |
|---|---|---|---|---|
| SI-NI-TI | 54.7 | 52.3 | 34.4 | 47.1 |
| SI-AI-TI (Ours) | **56.3** | **54.4** | **39.4** | **50.0** |
| SI-NI-DI | 39.5 | 37.2 | 19.3 | 32.0 |
| SI-AI-DI (Ours) | **50.8** | **51.5** | **27.7** | **43.3** |
| SI-NI-TI-DI | 62.3 | 59.3 | 41.7 | 54.4 |
| SI-AI-TI-DI (Ours) | **72.6** | **69.8** | **53.0** | **65.1** |

Furthermore, we generated adversarial examples on the ensemble models by using our SI-AI-TI-DI. As shown in Table 4, we achieved an average attack success rate of 95.0% on adversarially trained models under the black-box setting, which raised a new security issue for the robust deep neural networks.

**Table 4.** Comparison of SI-AI-TI-DI and SI-NI-TI-DI under ensemble-model setting. The adversarial examples are generated on the ensemble of Inc-v3, Inc-v4, IncRes-v2, and Res-101.

| Attack | Inc-v3$_{ens3}$ | Inc-v3$_{ens4}$ | IncRes-v2$_{ens}$ | Average |
|---|---|---|---|---|
| SI-NI-TI-DI | 95.4 | 93.7 | 89.5 | 92.9 |
| **SI-AI-TI-DI (Ours)** | **96.2** | **96.0** | **92.7** | **95.0** |

## 5. Discussion

It is commonly acknowledged that the training of neural network models is similar to the generation of adversarial examples, especially for gradient-based generation methods. Hence, techniques used in the training of neural networks, to improve model generalizability, can also be adopted to improve the transferability of adversarial examples. Since the Adam optimizer is often used in the training of neural networks, to improve convergence and achieve better performance on test sets, a second momentum term and decay step size in Adam were included in the AI-FGM algorithm. This was done to improve the transferability of adversarial examples. Furthermore, we suggest that other techniques (such as data augmentation) could be used to further improve the performance of adversarial examples in black-box environments.

## 6. Conclusions

In this study, we proposed the Adam Iterative Fast Gradient Method to improve the transferability of adversarial examples. Specifically, the Adam algorithm was modified to increase its suitability for the generation of adversarial examples. The proposed method improved both the iteration update direction and step size. The effectiveness of the proposed method was verified by an extensive series of experiments with ImageNet. Compared with previous gradient-based adversarial example generation techniques, our method improved attack success rates in black-box environments. In addition, we further improved the transferability of adversarial examples by combining our approach with existing methods and attacking ensemble models, which achieved a state-of-the-art attack success rate against adversarially trained networks. We suggest the proposed method could be used as a reference for other iterative gradient-based methods. For example, data augmentation could be combined with AI-FGM to achieve better attack performance.

**Data Availability**

The public data (dataset and models) can be downloaded from https://github.com/tensorflow/cleverhans/tree/master/examples/nips17_adverarial_competition/dataset, https://github.com/tensorflow/models/tree/master/research/slim and https://github.com/tensorflow/models/tree/master/research/adv_imagenet_models. Source code of the proposed method in this paper is available at https://github.com/YinHeng121/Adam_attack.

**Acknowledgments**

This work was supported by the National Key Research and Development Program of China, grant number 2017YFB0801900.

**Conflicts of Interest**

The authors declare that there are no conflicts of interest regarding the publication of this paper.

**Authors' Contributions**

Heng Yin and Hengwei Zhang contributed equally to this work.